\documentclass{article}

\usepackage[preprint,nonatbib]{blueprism}

\usepackage{cite}
\usepackage{amsmath,amssymb,amsfonts}
\usepackage{graphicx}
\usepackage{textcomp}
\usepackage{xcolor}

\usepackage{url}
\usepackage{booktabs}
\usepackage{multirow}
\usepackage{enumitem}
\usepackage[colorlinks]{hyperref}
\usepackage[capitalise]{cleveref}
\usepackage{microtype}

\usepackage{subcaption}

\usepackage[ruled,vlined,linesnumbered]{algorithm2e}
\usepackage{framed}

\usepackage{amsthm}
\theoremstyle{definition}

\usepackage{IEEEtrantools}

\usepackage[T1]{fontenc}

\begin{document}

\title{Process Discovery for Structured Program Synthesis}

\author{%
	Dell Zhang\thanks{Dell Zhang is the corresponding author of this paper. He is on leave from Birkbeck, University of London, and works full-time for Blue Prism AI Labs.}, Alexander Kuhnle, Julian Richardson, Murat Sensoy\\
	\texttt{dell.z@ieee.org}\\
	\texttt{\{alexander.kuhnle,julian.richardson,murat.sensoy\}@blueprism.com}
    \\\\
    Blue Prism AI Labs\\
}

\maketitle

\begin{abstract}
A core task in process mining is process discovery which aims to learn an accurate process model from event log data. 
In this paper, we propose to use (block-) structured programs directly as target process models so as to establish connections to the field of program synthesis and facilitate the translation from abstract process models to executable processes, e.g., for robotic process automation.
Furthermore, we develop a novel bottom-up agglomerative approach to the discovery of such structured program process models.
In comparison with the popular top-down recursive inductive miner, our proposed agglomerative miner enjoys the similar theoretical guarantee to produce sound process models (without deadlocks and other anomalies) while exhibiting some advantages like avoiding silent activities and accommodating duplicate activities.
The proposed algorithm works by iteratively applying a few graph rewriting rules to the directly-follows-graph of activities.
For real-world (sparse) directly-follows-graphs, the algorithm has quadratic computational complexity with respect to the number of distinct activities. 
To our knowledge, this is the first process discovery algorithm that is made for the purpose of program synthesis.
Experiments on the BPI-Challenge 2020 dataset and the Karel programming dataset have demonstrated that our proposed algorithm can outperform the inductive miner not only according to the traditional process discovery metrics but also in terms of the effectiveness in finding out the true underlying structured program from a small number of its execution traces.
\end{abstract}


\section{Introduction}
\label{sec:Introduction}

In recent years, there has been a surge of interest in \emph{process mining}~\cite{aalstProcessMiningData2016} and \emph{robotic process automation}~\cite{vanderaalstRoboticProcessAutomation2018}.
While the former addresses the problem of analyzing and optimizing processes, the latter tries to automate those mundane and repetitive tasks with software agents.
In the vision of \emph{hyperautomation}, predicted by Gartner in 2020 as the No. 1 strategic technology trend, advanced technologies including machine learning, process mining and robotic process automation need to be combined and coordinated in order to reap the full benefit of a digital workforce.

One area where process mining could be utilized to help robotic process automation is \emph{demo-to-process}~\cite{zhangProceedingsAAAI20Workshop2020,ferreiraEvaluationIntelligentProcess2020}, i.e., learning an executable software process from human demonstrations or \emph{user-interaction logs}~\cite{lenoRoboticProcessMining2020}.
This could relieve developers of the manual effort to build processes for robotic process automation.
However, although process mining and robotic process automation are widely regarded as ``a perfect match''~\cite{geyer-klingebergProcessMiningRobotic2018}, it is not straightforward to seamlessly integrate them together.
There is actually a gap between the process models discovered by today's process mining techniques and the process models ready to be deployed for robotic process automation.
For example, Petri net, which is probably the most popular process model used in process discovery, is alien to the mainstream robotic process automation systems such as Blue Prism, UiPath and Automation Anywhere.
It may be because existing process discovery techniques were designed for the processes of high-level business activities where complex phenomena like concurrency often occur and need to be captured, but for robotic process automation, the processes are of low-level user-interaction activities that must be able to be carried out by computer software. 

In fact, the processes in robotic process automation, at least of today, are essentially (block-) \emph{structured programs} that consist of simple control flow constructs.
Here ``structured'' has the same meaning as in the term ``structured programming'' coined by the computer scientist Edsger Dijkstra. 
The well-known structured program theorem (aka the B\"ohm–Jacopini theorem)~\cite{bohmFlowDiagramsTuring1966} tells us that any computable function can be represented using three control flow constructs --- sequence, selection and iteration --- as the only building blocks.
Why don't we use structured programs directly as the target process models for process discovery? 
Thus the discovered process models, i.e., the structured programs, can be fed straightaway to robotic process automation systems without any friction.
Driven by the above motivation, we have developed a new process discovery algorithm that learns structured programs directly from event logs.
The typical demo-to-process scenario is that a relatively simple process model (i.e., a short structured program) needs to be inferred from only a small number of demonstrations, which is quite different from the traditional process discovery problem setting where a relatively complex process model needs to be inferred from a large number of traces in the event log.  

The existing process discovery technique most similar to what we propose in this paper is the \emph{inductive miner}~\cite{leemansDiscoveringBlockStructuredProcess2013,leemansDiscoveringBlockStructuredProcess2013a,leemansDiscoveringBlockStructuredProcess2014,leemansScalableProcessDiscovery2015} which has \emph{process trees} as its target process models.  
Although structured programs, or equivalently their \emph{abstract syntax trees}, have the same expressive power as process trees, there are several nontrivial differences between them which make structured programs easier to understand and implement.  
Moreover, while the inductive miner recursively splits the \emph{directly-follows-graph}~\cite{leemansDiscoveringBlockStructuredProcess2013,leemansDirectlyFollowsbasedProcess2019} --- a graph that indicates what activities occurred right after what activities in the given event log --- in a \emph{top-down} fashion, our proposed process discovery algorithm works the other way around: it iteratively ``condenses'' the directly-follows-graph of activities \emph{bottom-up}.
That is why we name this approach \emph{agglomerative process discovery}.
In theory, there should be no fundamental difference between the hierarchical process models constructed top-down or bottom-up. 
However, in practice, we have found that the bottom-up approach is likely to generate better hierarchical process models than the top-down approach, probably because it is a lot easier to recognize local control flow constructs than global control flow constructs from the directly-follows-graph, as we will explain later.

For any input event log, our proposed process discovery algorithm finally outputs a program.
Therefore, it can also be considered as a method for \emph{program synthesis}~\cite{gulwaniProgramSynthesis2017}, or more specifically, \emph{programming by demonstration}~\cite{cypherWatchWhatProgramming1993,liebermanYourWishMy2001}.  
Following the steps of some recent work in neural program synthesis~\cite{devlinNeuralProgramMetaInduction2017,bunelLeveragingGrammarReinforcement2018,chenExecutionGuidedNeuralProgram2018,shinImprovingNeuralProgram2018}, we use Karel, a simple educational programming language~\cite{pattisKarelRobotGentle1981}, as the testbed to evaluate our proposed agglomerative miner and compare it with the inductive miner for the purpose of structured program synthesis. 
The experimental results on large-scale public datasets are encouraging.

\section{Related Work}
\label{sec:Related-Work}

\subsection{Process Discovery}
\label{sec:Process-Discovery}

One of the most important and most studied problems in process mining~\cite{aalstProcessMiningData2016} is process discovery, which tries to find a suitable process model to describe the control flow relations between the activities observed in or implied by a given event log~\cite{augustoAutomatedDiscoveryProcess2018}.
It is straightforward, but not really useful, to produce a process model that matches only the observed traces in the given event log (i.e., 100\% precision) or a process model that matches every possible trace (i.e., 100\% recall).
The central challenge for process discovery is to make the right trade-off and strike the optimal balance between \emph{precision}, \emph{recall} (more commonly known as \emph{fitness} in the process mining literature), \emph{generalization} and \emph{simplicity}~\cite{aalstProcessMiningData2016}. 

A well-known classic process discovery algorithm is the \emph{alpha miner} (the $\alpha$ algorithm)~\cite{vanderaalstWorkflowMiningDiscovering2004}.
It is able to find a Petri net model to fit the event log where all the activities are visible and unique.
One notable weakness of the alpha miner and many other process discovery algorithms is that the discovered model may not be sound, i.e., the model could suffer from anomalies like deadlocks.  

The most popular process discovery algorithm today is probably the \emph{inductive miner}\cite{leemansDiscoveringBlockStructuredProcess2013,leemansDiscoveringBlockStructuredProcess2013a,leemansDiscoveringBlockStructuredProcess2014}, especially its latest version based on directly-follows-graphs called IMD~\cite{leemansScalableProcessDiscovery2015}.
It produces a (block-) structured hierarchical process tree as the output model.
All the process trees are guaranteed to be sound, which might be the biggest strength of the inductive miner.
The basic idea of the inductive miner is to recursively detect an appropriate ``cut'' to split the directly-follows-graph~\cite{leemansDiscoveringBlockStructuredProcess2013,leemansDirectlyFollowsbasedProcess2019} top-down until the graph is divided into just individual activities (base cases).
There are four possible types of cuts: sequence, exclusive-choice, redo-loop and parallel. 
Often the process tree has to introduce silent/hidden activities ($\tau$) to capture the control flow, and it prohibits the existence of any duplicate activity.

\subsection{Program Synthesis}
\label{sec:Program-Synthesis}

The task of \emph{program synthesis}~\cite{gulwaniProgramSynthesis2017} is to automatically construct a program (in the underlying programming language) that can satisfy a user intent expressed in some form of high-level specification.
This sub-field of AI has a long history and it has been considered as the ``holy grail'' of computer science.
In recent years, it has attracted a lot of attention due to the popularization of practical program synthesis applications (like the FlashFill feature in Microsoft Excel~\cite{gulwaniAutomatingStringProcessing2011,gulwaniSpreadsheetDataManipulation2012}) and also the great potential of deep learning for neural program synthesis~\cite{devlinNeuralProgramMetaInduction2017,bunelLeveragingGrammarReinforcement2018}.
Popular program synthesis frameworks include \textsc{PROSE}, \textsc{Sketch}, \textsc{Rosette} and \textsc{Foofah}. 

Most of the recent work in this area aims to learn simple programs (e.g., in the Karel programming language~\cite{pattisKarelRobotGentle1981}) only from input-output examples or, in a couple of recent studies~\cite{chenExecutionGuidedNeuralProgram2018,shinImprovingNeuralProgram2018}, by additionally exploiting the (inferred) execution traces to improve program synthesis.
In process mining, we usually instead assume the availability of traces (e.g., stored in an event log) but not input-output examples. 

\section{Proposed Approach}
\label{sec:Proposed-Approach}

\subsection{Structured Program}
\label{sec:Structured-Program}

The target process model for our proposed approach to process discovery is just (block-) structured programs that are formally defined in \cref{tab:structured_program}.
The alphabet $\Sigma$ is the finite set of activities that can occur in the event log, and the sole non-terminal symbol $S$ represents a structured program.
As shown by the production rules, $S$ can be either a simple statement which consists of a single activity (terminal symbol), or a compound statement that is made from a control flow construct with smaller program pieces as its components.

To denote the three standard control flow constructs in structured programming, i.e., sequence, selection and iteration, we borrow the widely used \emph{regular expression} operators. 
Specifically, 
the $?$ operator indicates an optional occurrence of its preceding statement $S$;
the $|$ operator indicates the occurrence of either the statement on its left $S_1$ or the statement on its right $S_2$;
the $+$ operator indicates one  or more occurrences of its preceding statement $S$; and 
the $*$ operator indicates zero or more occurrences of its preceding statement $S$.
Parentheses are used to group statements for the application of operators.

In addition to the above standard control flow constructs, we also include the concurrence construct (with the $\&$ operator) in order to represent the parallel execution of statements.
This is necessary to make our process model comparable with the other process models in process discovery.
Some programming languages like Erlang have built-in primitives to support concurrent or parallel computing.
For the programming languages without this capability, such as Python, it can be implemented via an add-on library (e.g., \texttt{multiprocessing}) or simply backing off to the serialized execution of statements. 

\begin{table*}[!tb]
	\caption{The constituency grammar of structured programs.}
	\label{tab:structured_program}
	\centering
	\begin{tabular}{@{}l|l|l@{}}
		\toprule
		Production Rule & Control Flow & Source Code \\
		\midrule
		$S \longrightarrow x$ & activity $x \in \Sigma$ & 
			\begin{minipage}{0.2\columnwidth}
				$x$
			\end{minipage} \\
		\midrule
		$S \longrightarrow (S_1 S_2)$ & sequence  & 
			\begin{minipage}{0.2\columnwidth}
				$S_1$ \\
				$S_2$
			\end{minipage} \\	
		\midrule
		$S \longrightarrow (S_1?)$ & selection & 
			\begin{minipage}{0.2\columnwidth}
				\verb|if |(.): \\
				\verb|  |$S_1$
			\end{minipage} \\
		\midrule
		$S \longrightarrow (S_1 | S_2)$ & selection & 
			\begin{minipage}{0.2\columnwidth}
				\verb|if |(.): \\
				\verb|  |$S_1$ \\
				\verb|else|: \\
				\verb|  |$S_2$ 
			\end{minipage} \\	
		\midrule
		$S \longrightarrow (S_1+)$ & iteration & 
			\begin{minipage}{0.2\columnwidth}
				\verb|while |(.): \\
				\verb|  |$S_1$
			\end{minipage} \\
		\midrule
		$S \longrightarrow (S_1*)$ & iteration & 
			\begin{minipage}{0.2\columnwidth}
				\verb|while |(.): \\
				\verb|  |$S_1$
			\end{minipage} \\
		\midrule
		$S \longrightarrow (S_1 \& S_2)$ & concurrence & 
			\begin{minipage}{0.2\columnwidth}
				\verb|para|: \\
				\verb|  |$S_1$ \\
				\verb|  |$S_2$ 
			\end{minipage} \\
		\bottomrule
	\end{tabular}
\end{table*}

\begin{figure*}[!tb]
    \centering
    \begin{subfigure}[T]{0.6\columnwidth}
    	\centering
		\includegraphics[scale=0.36]{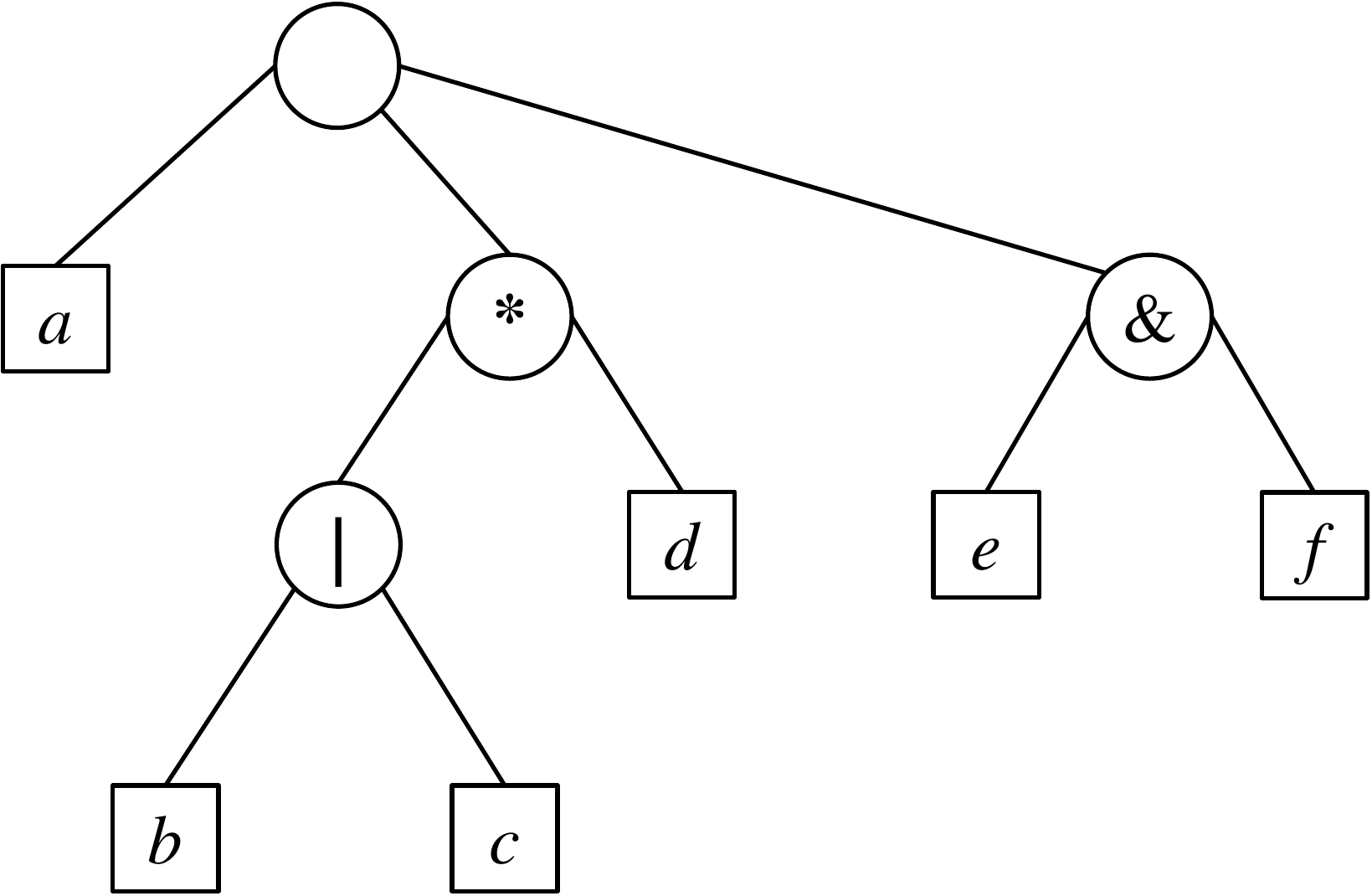}
		\bigskip
        \caption{abstract syntax tree}
		\label{fig:abstract_syntax_tree}
    \end{subfigure}
    \hfill    
    \begin{subfigure}[T]{0.3\columnwidth}
    	\centering
		\begin{minipage}{\textwidth}
    	\begin{framed}
			$a$ \\
			\verb|while |(.): \\
			\verb|  if |(.): \\
			\verb|    |$b$ \\
			\verb|  else|: \\
			\verb|    |$c$ \\
			\verb|  |$d$ \\
			\verb|para|: \\
			\verb|  |$e$ \\			
			\verb|  |$f$ 			
		\end{framed}
		\end{minipage}
        \caption{source code}
		\label{fig:source_code}
    \end{subfigure}
    \caption{An example structured program $( a ( (b|c) d)\text{*}(e\&f) )$.}
	\label{fig:structured_program}
\end{figure*}

Such a structured program can be represented equivalently by its \emph{abstract syntax tree}, where the leaf nodes are activities (simple statements) and the internal nodes are operators (to construct compound statements), as illustrated in \cref{fig:structured_program}.

The (structured program) syntax tree model looks very similar to the \emph{process tree} model used by the inductive miner~\cite{leemansDiscoveringBlockStructuredProcess2013}.
Indeed, they should have the same expressive power, and they are both block-structured process models that are \emph{sound} by construction.
However, syntax trees are tailored towards program synthesis and differ from process trees in two important aspects.
First, syntax trees completely avoid the usage of invisible \emph{silent activities} ($\tau$) which are unintuitive.
Second, syntax trees describe iterations with the standard concept of while-loops (as in almost all programming languages), instead of the obscure \emph{redo-loops} which consists of not only a ``do'' part but also one or more ``re-do'' parts.
Therefore, syntax trees are easier to interpret and implement. 

To a large degree, the (structured program) syntax tree model also resembles \emph{regular expressions} (defined by a \emph{regular language}), except that syntax trees can also model concurrency with the additional $\&$ operator.
A well-known theorem in computer science established by E Mark Gold states that even regular expressions cannot be \emph{learned in the limit} from positive examples only~\cite{goldLanguageIdentificationLimit1967}, though the problem of \emph{inductive inference} has been investigated for a variety of subclasses~\cite{angluinInductiveInferenceTheory1983}.
Most existing process discovery algorithms seem to overcome this obstacle to learn from positive examples (observed traces) only by imposing a strong \emph{inductive bias} against duplicate activities in the process model.

\subsection{Agglomerative Miner}
\label{sec:Agglomerative-Miner}

Our proposed agglomerative approach to process discovery is given in Algorithm~\ref{alg:agglomerative}.
Let us explain it in detail and compare it with the inductive miner~\cite{leemansScalableProcessDiscovery2015}.

\begin{algorithm*}[!tb]
	\SetKwInOut{Input}{Input}
	\SetKwInOut{Output}{Output}
	\SetAlgoLined
	\Input{An event log $L$.}
	\Output{A structured program $S$.}
	\BlankLine
	\For{each trace $\langle{a_1,\dots,a_k}\rangle \in L$}{
		Expand it to $\langle\wedge,{a_1,\dots,a_k},$\$$\rangle$ where $\wedge$ and \$ are the special `begin' and `end' activities respectively\;
	}
	Construct the directly-follows-graph $G$ for $S$\;
	\While{not converged}{
		\Repeat{$G$ cannot be condensed further}{
			Condense $G$ using the graph rewriting rule 
			\cref{fig:iteration1} iteration1 (self-loop)\;
		}
		\Repeat{$G$ cannot be condensed further}{
			Condense $G$ using the graph rewriting rule 
			\cref{fig:sequence} sequence\;
		}
		\Repeat{$G$ cannot be condensed further}{
			Condense $G$ using the graph rewriting rules 
			\cref{fig:iteration2,fig:iteration3,fig:iteration4,fig:iteration5,fig:iteration6} iteration2-6 (general-loop) as well as \cref{fig:concurrence} concurrence\;
		}
		\Repeat{$G$ cannot be condensed further}{
			Condense $G$ using the graph rewriting rule 
			\cref{fig:selection1}: selection1 (multi-branch)\;
		}
		\Repeat{$G$ cannot be condensed further}{
			Condense $G$ using the graph rewriting rule 
			\cref{fig:selection2}: selection2 (single-branch)\;
		}
	}
	\If{$G$ contains more than one node other than $\wedge$ and \$}{
		Condense $G$ using the fall-through ``flower'' model as the last resort\;
	}
	$S$ $\longleftarrow$ the structured program saved at the node $v$, the only node left other than $\wedge$ and \$\;
	\KwRet{$S$}
	\caption{Agglomerative Process Discovery}
	\label{alg:agglomerative}
\end{algorithm*}

As with all existing process discovery algorithms, the agglomerative miner takes an event log $L$ as the input and produces a process model $S$ as the output. 
Here, the input event log $L$ is a bag (multiset) of \emph{traces}, each of which consists of a sequence of activities, and the output process model $S$ will be a structured program (or equivalently its abstract syntax tree).

\begin{figure*}
	\centering
	\includegraphics[scale=0.36]{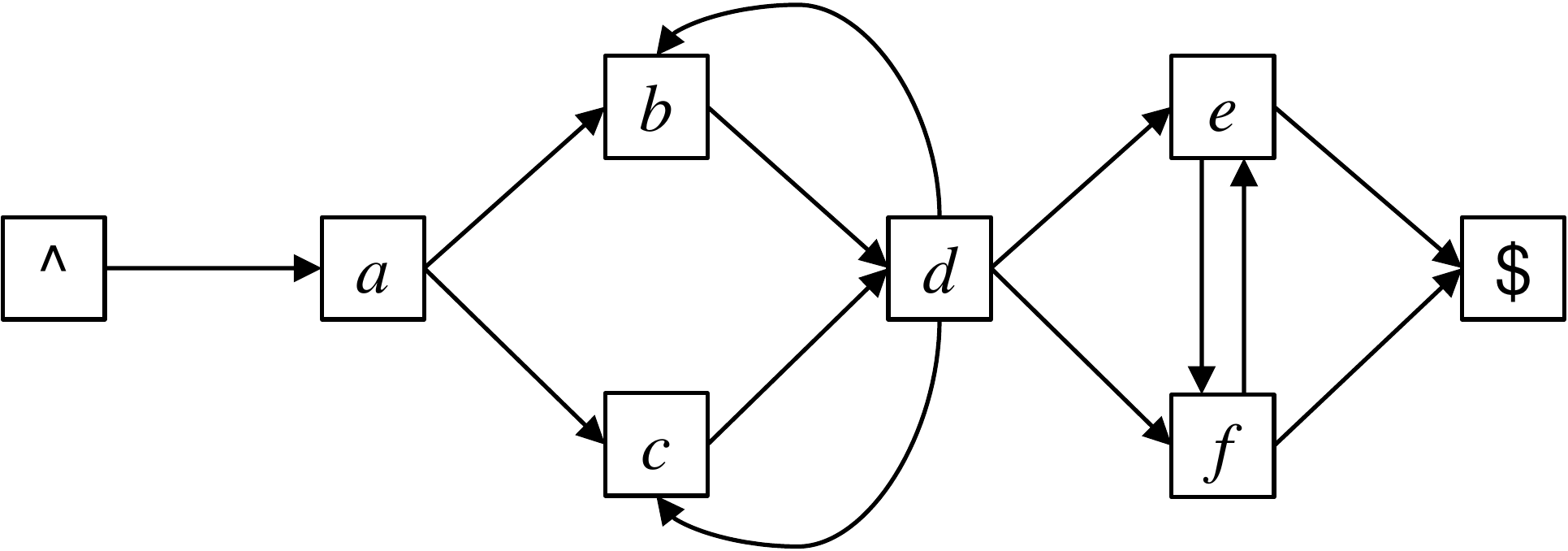}
	\caption{The directly-follows-graph constructed from the execution traces of the example structured program.}
	\label{fig:directly-follows-graph}
\end{figure*}

Similar to the inductive miner, the agglomerative miner first converts the given event log into a \emph{directly-follows-graph} which has a directed edge (link) from a node (activity) $u$ to another node (activity) $v$, if and only if, $u$ is directly followed by $v$.
Unlike the inductive miner, the agglomerative miner does not need to memorize the start and end activities of the directly-follows-graph. 
Instead, we introduce two special activities $\wedge$ and \$ to represent the beginning and end of traces respectively, which simplifies the algorithm.
\cref{fig:directly-follows-graph} shows the directly-follows-graph corresponding to the example structured program in \cref{fig:structured_program}.

The body of the agglomerative process discovery algorithm is an iterative procedure of graph rewriting that condenses the directly-follows-graph step by step until only one node (other than $\wedge$ and \$) is left.
Along with this iterative procedure, the activities represented by the nodes of the directly-follows-graph are pieced together through different control flow constructs gradually into a complete structured program, which is the final inferred process model.
The overall framework in which the input graph is summarized into a single node containing the output bears some similarity to the \emph{state elimination} method for transforming nondeterministic finite automata into regular expressions~\cite{hopcroftIntroductionAutomataTheory2007,grossHandbookGraphTheory2013} that can be traced back to Kleene~\cite{kleeneRepresentationEventsNerve1956}.

\begin{figure*}[!tb]
	\centering
	\begin{minipage}{\linewidth}
	\begin{framed}
    \begin{subfigure}[T]{0.48\linewidth}
    	\centering
        \includegraphics[scale=0.36]{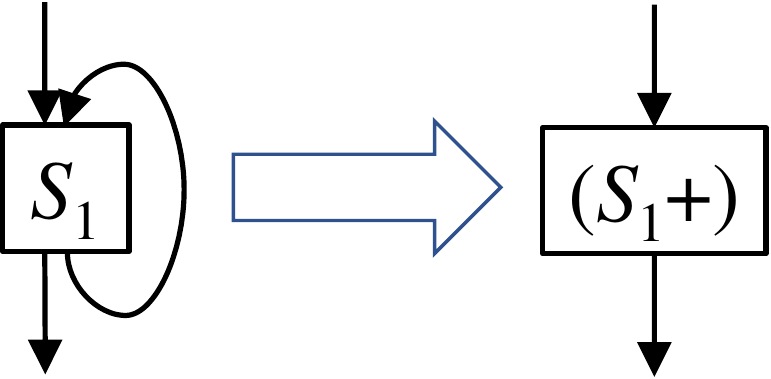}
        \caption{iteration1}
        \label{fig:iteration1}
    \end{subfigure}
    \hspace{0.04\linewidth}
    \begin{subfigure}[T]{0.48\linewidth}
    	\centering
        \includegraphics[scale=0.36]{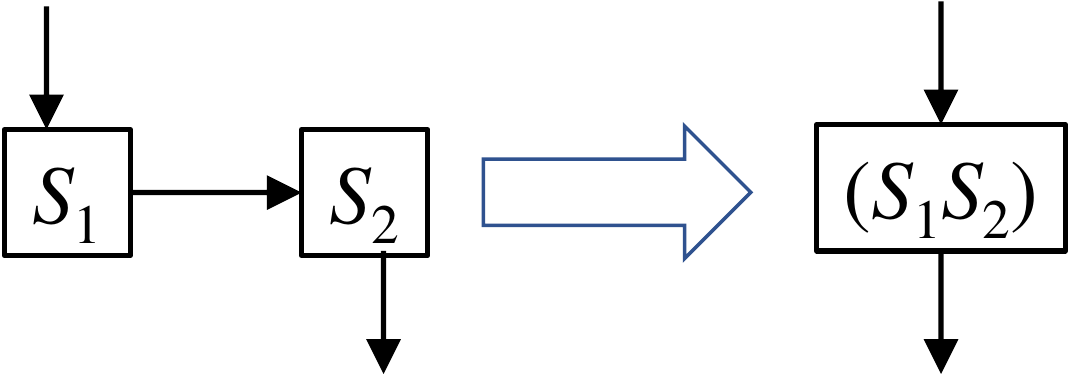}
        \caption{sequence}
        \label{fig:sequence}
    \end{subfigure}
    \smallskip \\
    \begin{subfigure}[T]{0.48\linewidth}
    	\centering
        \includegraphics[scale=0.36]{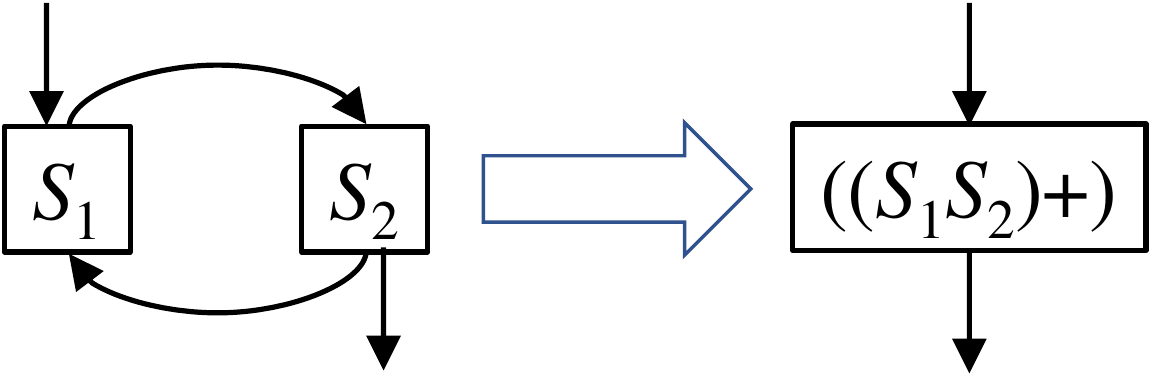}
        \caption{iteration2}
        \label{fig:iteration2}
    \end{subfigure}
    \hspace{0.04\linewidth}
    \begin{subfigure}[T]{0.48\linewidth}
    	\centering
        \includegraphics[scale=0.36]{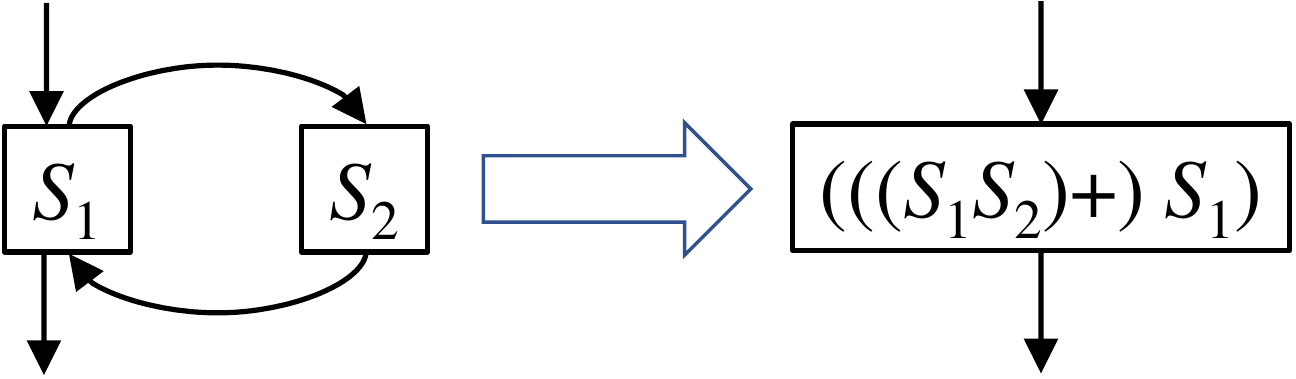}
        \caption{iteration3}
        \label{fig:iteration3}
    \end{subfigure}
    \smallskip \\ 
    \begin{subfigure}[T]{0.48\linewidth}
    	\centering
        \includegraphics[scale=0.36]{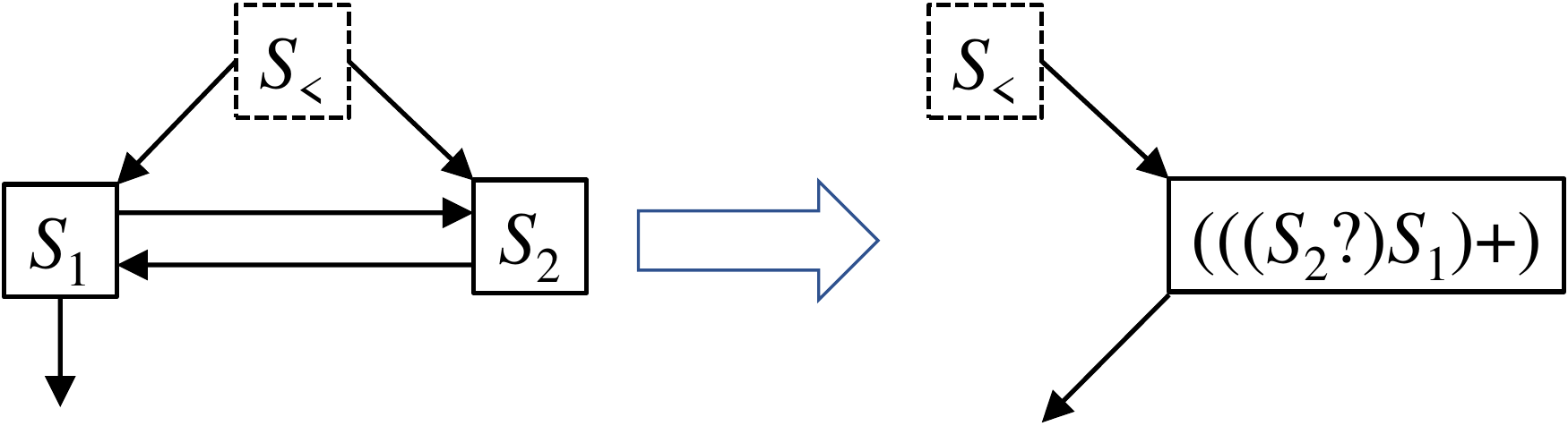}
        \caption{iteration4}
        \label{fig:iteration4}
    \end{subfigure}
    \hspace{0.04\linewidth}
    \begin{subfigure}[T]{0.48\linewidth}
    	\centering
        \includegraphics[scale=0.36]{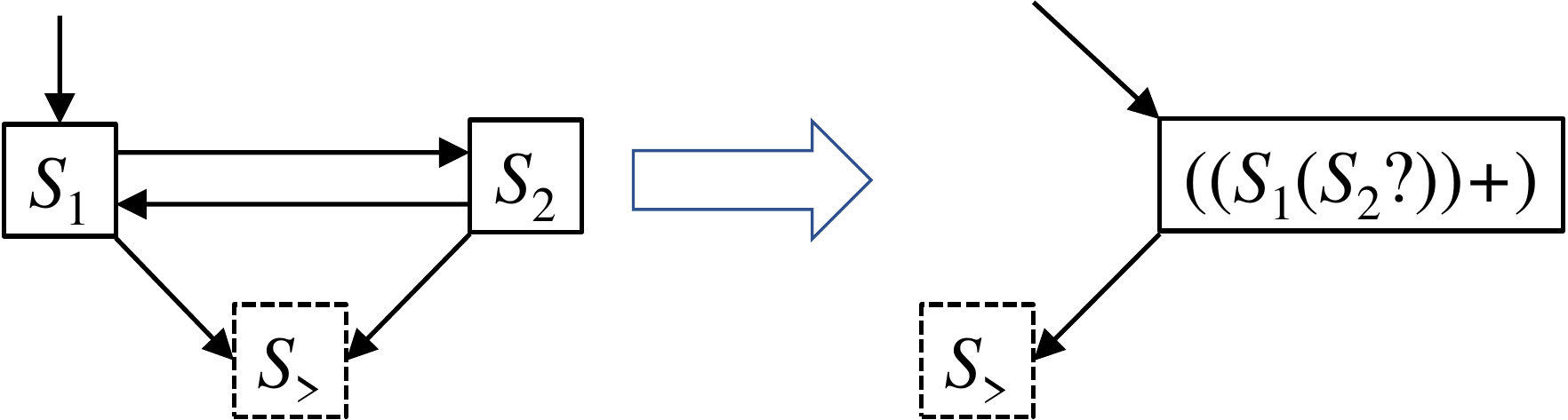}
        \caption{iteration5}
        \label{fig:iteration5}
    \end{subfigure}
    \smallskip \\
    \begin{subfigure}[T]{0.48\linewidth}
    	\centering
        \includegraphics[scale=0.36]{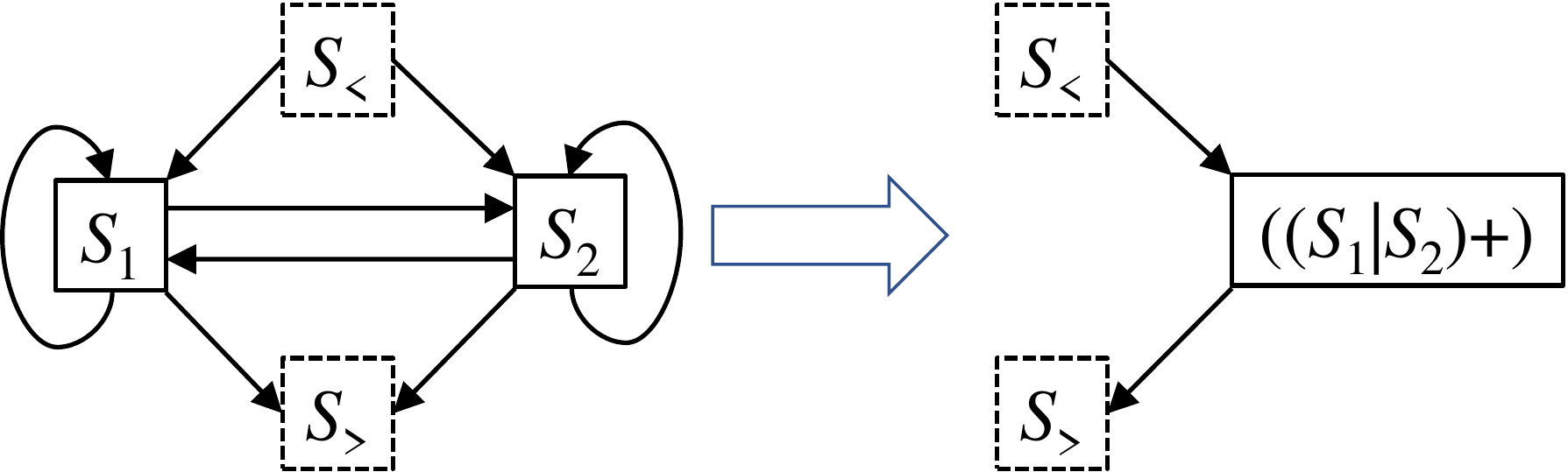}
        \caption{iteration6}
        \label{fig:iteration6}
    \end{subfigure}
    \hspace{0.04\linewidth}
    \begin{subfigure}[T]{0.48\linewidth}
    	\centering
        \includegraphics[scale=0.36]{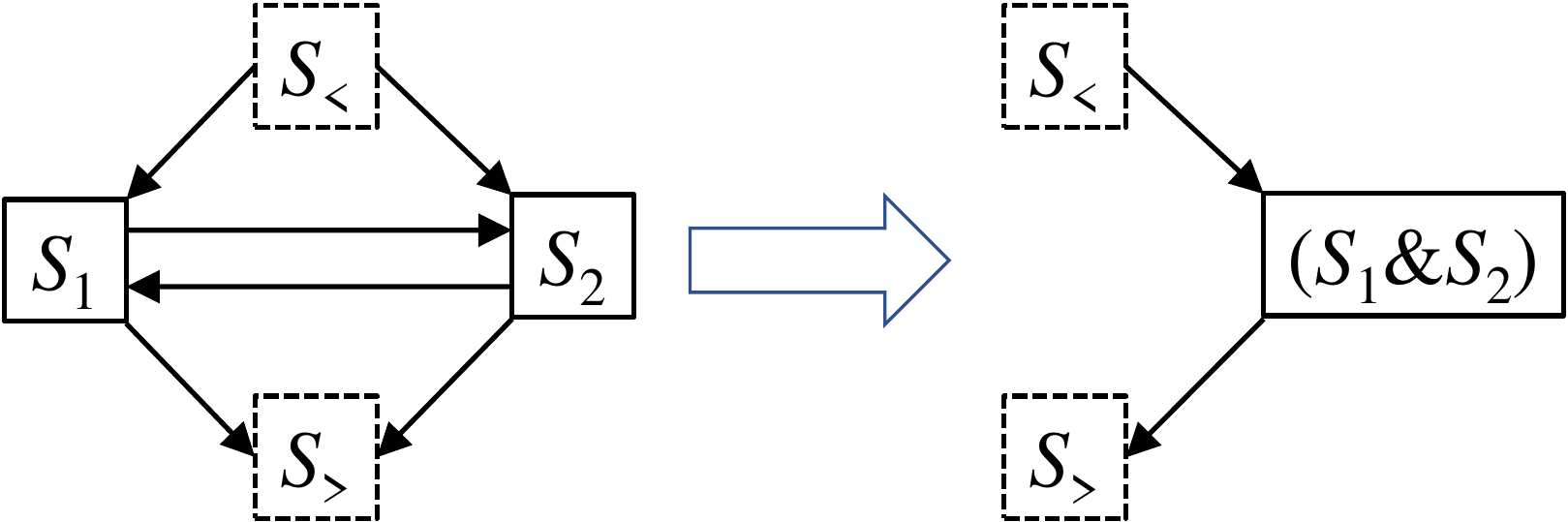}
        \caption{concurrence}
        \label{fig:concurrence}
    \end{subfigure}
    \smallskip \\ 
    \begin{subfigure}[T]{0.48\linewidth}
    	\centering
        \includegraphics[scale=0.36]{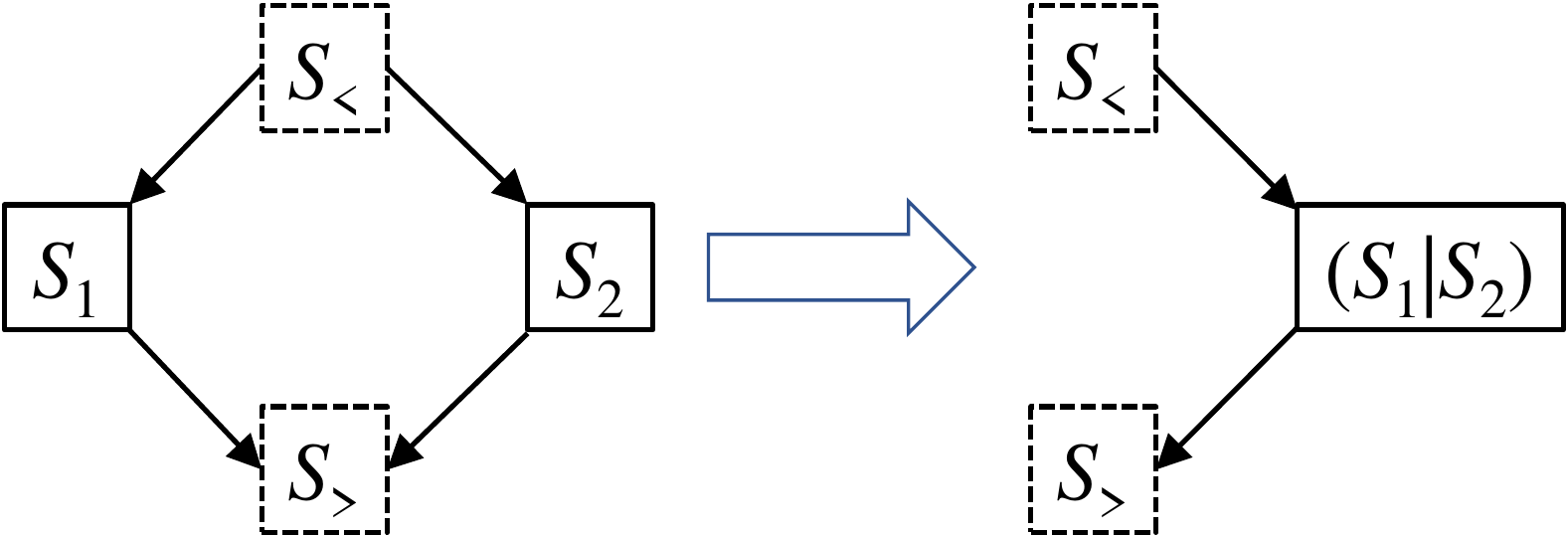}
        \caption{selection1}
        \label{fig:selection1}
    \end{subfigure}
    \hspace{0.04\linewidth}
    \begin{subfigure}[T]{0.48\linewidth}
    	\centering
        \includegraphics[scale=0.36]{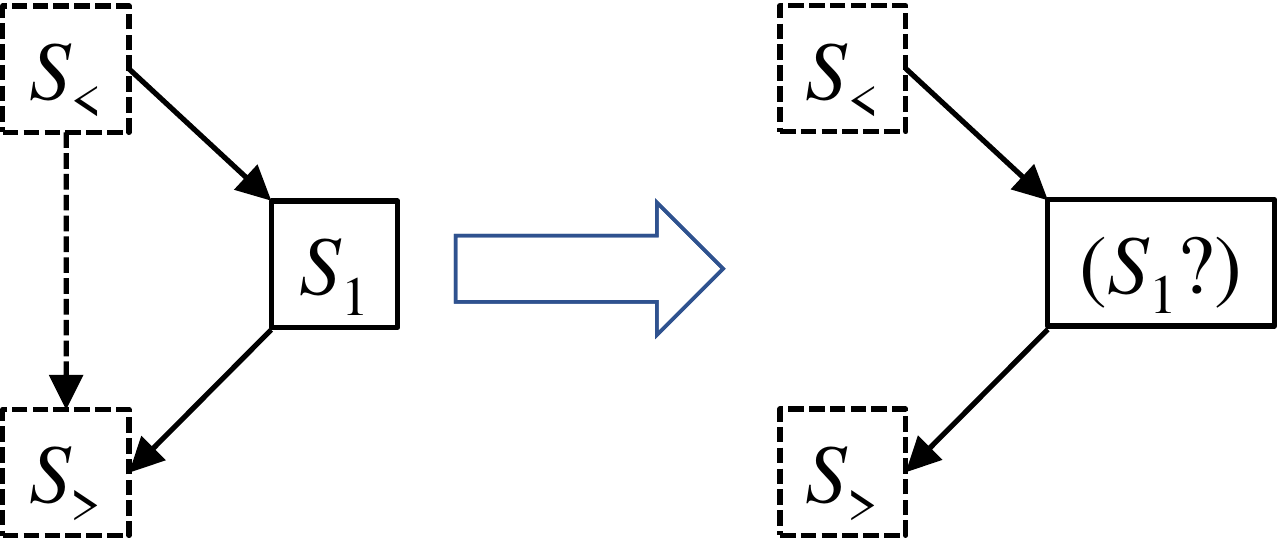}
        \caption{selection2}
        \label{fig:selection2}
    \end{subfigure}
    \end{framed}
    \end{minipage}
	\caption{The graph rewriting rules for the agglomerative process discovery algorithm.}
	\label{fig:graph rewriting}
\end{figure*}

\cref{fig:graph rewriting} shows all the graph rewriting rules employed by the agglomerative miner.
Each of them detects a particular local graph pattern and then uses the corresponding control flow construct to condense the graph.
For example, as shown in \cref{fig:sequence}, if we see that a node $S_1$ has only one successor $S_2$ and the node $S_2$ has only one predecessor $S_1$, we can safely merge $S_1$ and $S_2$ into a sequence $(S_1 S_2)$.
Those graph rewriting rules cover all the basic local graph patterns involving either one or two nodes.
The order in which they are applied matters: the general principle here is that the graph rewriting rules with less ambiguity should be applied before those with more ambiguity.
Since there is no uncertainty about the self-loop pattern \cref{fig:iteration1} and the sequence pattern \cref{fig:sequence}, these are always attempted first at each iteration of the algorithm. 

One particular graph rewriting rule, ``iteration3'' (\cref{fig:iteration3}), is of particular importance, as it leads to a piece of structured program $(((S_1 S_2)+)S_1)$ which contains the duplication of statement $S_1$.
This distinguishes the agglomerative miner from most existing process discovery algorithms including the inductive miner, as they do not accommodate \emph{duplicate activities} in the process model. 
When $S_1$ is a simple statement (containing just an activity), duplicating $S_1$ is preferable because it results in a relatively small process model that can faithfully fit this local graph pattern.
However, when $S_1$ is a compound statement (containing multiple activities), duplicating $S_1$ would make the generated process model much bigger which is probably undesirable, so we may actually want to sacrifice precision for simplicity and resolve to $((S_1(S_2?))+)$ instead.

Each time we utilize a graph rewriting rule to condense the directly-follows-graph, we also try to simplify the piece of structured program to be produced. 
\cref{tab:simiplification} lists the simplification rules, each of which converts the given structured program into a semantically equivalent but syntactically simpler one.
For example, if the final operator used by a graph rewriting rule is `$?$', then the first three simplification rules would be applicable.

\begin{table*}[!tb]\small
	\caption{The simplification of structured programs.}
	\label{tab:simiplification}
	\centering
	\begin{tabular}{@{}lcl@{}}
		\toprule
		Original Expression  & $\Rightarrow$ & Simplified Expression \\
		\midrule
		$((S?)?)$            & $\Rightarrow$ & $(S?)$                \\
		$((S+)?)$            & $\Rightarrow$ & $(S*)$                \\
		$((S*)?)$            & $\Rightarrow$ & $(S*)$                \\
		$((S+)+)$            & $\Rightarrow$ & $(S+)$                \\
		$((S?)+)$            & $\Rightarrow$ & $(S*)$                \\
		$((S*)+)$            & $\Rightarrow$ & $(S*)$                \\
		$((S?)*)$            & $\Rightarrow$ & $(S*)$                \\
		$((S+)*)$            & $\Rightarrow$ & $(S*)$                \\
		$((S*)*)$            & $\Rightarrow$ & $(S*)$                \\
		$((S_1 S_2) S_3)$    & $\Rightarrow$ & $(S_1 S_2 S_3)$       \\
		$(S_1 (S_2 S_3))$    & $\Rightarrow$ & $(S_1 S_2 S_3)$       \\
		$((S_1|S_2)|S_3)$    & $\Rightarrow$ & $(S_1|S_2|S_3)$       \\
		$(S_1|(S_2|S_3))$    & $\Rightarrow$ & $(S_1|S_2|S_3)$       \\
		$((S_1\&S_2)\&S_3)$  & $\Rightarrow$ & $(S_1\&S_2\&S_3)$     \\
		$(S_1\&(S_2\&S_3))$  & $\Rightarrow$ & $(S_1\&S_2\&S_3)$     \\
		$((S_1?)|S_2)$       & $\Rightarrow$ & $((S_1|S_2)?)$        \\
		$(S_1|(S_2?))$       & $\Rightarrow$ & $((S_1|S_2)?)$        \\
		$(((S_1+)|S_2)+)$    & $\Rightarrow$ & $((S_1|S_2)+)$        \\
		$((S_1|(S_2+))+)$    & $\Rightarrow$ & $((S_1|S_2)+)$        \\
		$(((S_1*)|S_2)+)$    & $\Rightarrow$ & $((S_1|S_2)*)$        \\
		$((S_1|(S_2*))+)$    & $\Rightarrow$ & $((S_1|S_2)*)$        \\
		$(((S_1+)|S_2)*)$    & $\Rightarrow$ & $((S_1|S_2)*)$        \\
		$((S_1|(S_2+))*)$    & $\Rightarrow$ & $((S_1|S_2)*)$        \\
		$(((S_1*)|S_2)*)$    & $\Rightarrow$ & $((S_1|S_2)*)$        \\
		$((S_1|(S_2*))*)$    & $\Rightarrow$ & $((S_1|S_2)*)$        \\
		$(((S_1?)(S_2?))+)$  & $\Rightarrow$ & $((S_1|S_2)*)$        \\
		$(((S_1*)(S_2*))+)$  & $\Rightarrow$ & $((S_1|S_2)*)$        \\
		$(((S_1?)(S_2?))*)$  & $\Rightarrow$ & $((S_1|S_2)*)$        \\
		$(((S_1*)(S_2*))*)$  & $\Rightarrow$ & $((S_1|S_2)*)$        \\
		$(((S_1+)(S_2?))+)$  & $\Rightarrow$ & $((S_1(S_2?))+)$      \\
		$(((S_1?)(S_2+))+)$  & $\Rightarrow$ & $(((S_1?)S_2)+)$      \\
		$(((S_1+)(S_2?))*)$  & $\Rightarrow$ & $((S_1(S_2?))*)$      \\
		$(((S_1?)(S_2+))*)$  & $\Rightarrow$ & $(((S_1?)S_2)*)$      \\
		\bottomrule
	\end{tabular}
\end{table*}

The above described iterative graph rewriting procedure is guaranteed to converge, as the application of each graph rewriting rule reduces the directly-follows-graph by either eliminating a node (i.e., contracting a pair of nodes into one) or eliminating an edge. 
Therefore the number of iterations is bounded by the size of the graph, and in practice a few iterations are usually enough to reach convergence.

In case there are still more than one node (other than $\wedge$ and \$) after the iterative graph rewriting procedure has converged, we summarize those remaining intermediate nodes $S_1,\ldots,S_k$ using the fall-through ``flower'' model $((S_1|\ldots|S_k)+)$.
Since this so-called flower model can fit any trace of activities, it is used as the last resort by many process discovery algorithms including the inductive miner.

In some graph rewriting rules (\cref{fig:iteration4,fig:iteration5,fig:iteration6,fig:concurrence,fig:selection1,fig:selection2}), we require the two nodes involved have a common predecessor ($S_<$) or/and a common successor ($S_>$), which is to ensure that the generated control flow construct has a well-defined entry and exit point.
Moreover, for the graph rewriting rule \cref{fig:selection1} (selection with two branches), the above constraint also helps to reduce the computational complexity: without this constraint, detecting the corresponding local graph pattern would require us to enumerate all the possible node pairs; with this constraint, however, we only need to check the successors of each node, which is much cheaper computationally.

Overall, for an event log with $n$ distinct activities, the computational complexity of the agglomerative process discovery algorithm is $O(n^2)$ if the directly-follows-graph is sparse (which is likely to be true for real-world datasets), or $O(n^3)$ otherwise. 
The reasoning is as follows.
The construction of the directly-follows-graph can be easily done with a sequential scan of the event log, so its time cost is negligible.
The number of nodes in the directly-follows-graph $|V|$ is obviously just the number of distinct activities $n$.
The number of edges $|E|$ is at most $n^2$. 
For ``sparse'' graphs (as commonly defined in graph theory or network science), $|E|$ is at the level of $O(n)$ instead of $O(n^2)$. 
The most computationally expensive part of the algorithm, the graph rewriting procedure, has up to $|V|+|E|$ iterations, as explained earlier. 
It is not difficult to see that at each iteration, the application of all the graph rewriting rules requires at most $O(|V|+|E|)$ steps.
Consequently, the total computational complexity is $O((|V|+|E|)^2) = O(n^2)$ for sparse directly-follows-graphs.

In summary, our proposed agglomerative miner is inspired by the popular inductive miner -- especially its latest version based on directly-follows-graphs -- but it is different from the inductive approach in a range of nontrivial aspects:
it works bottom-up rather than top-down;
it relies on iteration rather than recursion;
it outputs syntax trees instead of process trees; and
it avoids silent activities but accommodates duplicate activities in the final process model.

The agglomerative and inductive miner are both guaranteed to produce \emph{sound} process models (without deadlocks and other anomalies).
Why is it possible for the bottom-up agglomerative approach to find better process models than the top-down inductive approach?
We conjecture that it is because the inductive miner often has to make hard choices among different possible control flow constructs at early stages. 
By contrast, the agglomerative miner starts from extracting the obvious (unambiguous) local graph patterns using fine-grained graph rewriting rules which simplify the graph, then in subsequent iterations previously complex (ambiguous) graph patterns become straightforward in the simplified graph and thus can be further collapsed; this iterative graph rewriting procedure continues until the entire directly-follows-graph is summarized into a single piece of structured program.
While the inductive miner employs only four global graph patterns for recursive graph splitting, the agglomerative miner defines ten local graph patterns (as listed in \cref{fig:graph rewriting}), and more could be added if necessary.    

\section{Experiments}
\label{sec:Experiments}

We have conducted experiments on two datasets, one in the traditional process discovery setting and the other for the purpose of program synthesis, to empirically evaluate the proposed agglomerative miner and compare it with existing process discovery methods, including the classic alpha miner and the popular inductive miner.
For the inductive miner, we are referring to its latest version based on directly-follows-graphs, IMD~\cite{leemansScalableProcessDiscovery2015}, as it is scalable and most similar to our proposed agglomerative miner.

The agglomerative miner is implemented in Python 3, and we take implementations of the alpha miner as well as the inductive miner (specifically IMDFc) from the open-source Python library PM4Py\footnote{\url{https://pm4py.fit.fraunhofer.de/}}.

\subsection{BPI-Challenge 2020}
\label{sec:BPI-Challenge-2020}

The BPI-Challenge 2020 dataset\footnote{\url{https://icpmconference.org/2020/bpi-challenge/}} is a newly released public benchmark dataset for process mining.
It contains five large-scale event logs pertaining to two years of travel expense claims at the Eindhoven University of Technology (TU/e).

We adopt the following standard performance metrics for automated process discovery which have been widely used in the process mining research literature~\cite{augustoAutomatedDiscoveryProcess2018}: fitness, precision~\cite{munoz-gamaFreshLookPrecision2010}, $F_1$-score, generalization~\cite{buijsQualityDimensionsProcess2014}, and simplicity~\cite{blumMetricsProcessDiscovery2015}.
Among them, $F_1$-score is the harmonic mean of fitness (i.e., recall) and precision which reflects the overall accuracy of process discovery.

\cref{tab:bpi20-dataset,tab:bpi20-results} show the dataset statistics and the experimental results of the three process mining algorithms in comparison. 
It can be clearly seen that the agglomerative miner achieves the best $F_1$-score as well as simplicity on all five event logs, and it is also the best-performing model with respect to generalization on three out of five event logs.
This confirms the effectiveness of our proposed agglomerative miner for traditional process discovery with many traces and complex models.

\begin{table*}[!tb]
	\caption{Descriptive statistics of the BPI-Challenge 2020 dataset.}
	\label{tab:bpi20-dataset}
	\centering
	\begin{tabular}{@{}l|rr@{}}
		\toprule
		Log & \#Cases & \#Events \\
		\midrule
		Domestic Declarations      & 10,500  &   56,437 \\
		International Declarations &  6,449  &   72,151 \\
		Prepaid Travel Cost        &  2,099  &   18,246 \\
		Request for Payment        &  6,886  &   36,796 \\
		Travel Permits             &  7,065  &   86,581 \\
		\bottomrule
	\end{tabular}
\end{table*}

\begin{table*}[!tb]
	\caption{Process discovery performances on the BPI-Challenge 2020 dataset.}
	\label{tab:bpi20-results}
	\centering
	\begin{tabular}{@{}l|l|ccccc@{}}
		\toprule
		Log & Method & Fitness & Precision & $F_1$-score & Generalization & Simplicity \\
		\midrule
		\multirow{3}{2cm}{Domestic Declarations}            &
		Alpha         &         0.7063  &         0.2500  &         0.3693  &         0.7851  &         0.1494  \\
		& 
		Inductive     &         0.9400  &         0.3086  &         0.4647  &         0.7540  &         0.5769  \\
		& 
		Agglomerative &         1.0000  &         0.3286  & \textbf{0.4946} & \textbf{0.8304} & \textbf{0.5814} \\
		\hline
		\multirow{3}{2cm}{International Declarations}       & 
		Alpha         &         0.5786  &         0.0000  &         0.0000  &         0.8987  &         0.1296  \\
		& 
		Inductive     &         1.0000  &         0.0958  &         0.1748  & \textbf{0.9029} &         0.5273  \\
		& 
		Agglomerative &         0.8279  &         0.2550  & \textbf{0.3899} &         0.9004  & \textbf{0.6048} \\
		\hline
		\multirow{3}{2cm}{Prepaid Travel Cost}              & 
		Alpha         &         0.6456  &         0.0000  &         0.0000  & \textbf{0.8773} &         0.0554  \\
		& 
		Inductive     &         0.9982  &         0.1111  &         0.2000  &         0.8700  &         0.5238  \\
		& 
		Agglomerative &         0.8294  &         0.1736  & \textbf{0.2871} &         0.8675  & \textbf{0.5942} \\
		\hline
		\multirow{3}{2cm}{Request for Payment}              & 
		Alpha         &         0.7610  &         0.0000  &         0.0000  &         0.7836  &         0.1396  \\
		& 
		Inductive     &         0.9317  &         0.2168  &         0.3517  &         0.8038  &         0.5842  \\
		& 
		Agglomerative &         0.9340  &         0.2284  & \textbf{0.3670} & \textbf{0.8805} & \textbf{0.6577} \\
		\hline
		\multirow{3}{2cm}{Travel Permits}                   &
		Alpha         &         0.5850  &         0.0000  &         0.0000  &         0.8553  &         0.1725  \\
		& 
		Inductive     &         0.9996  &         0.0708  &         0.1323  &         0.8110  &         0.4912  \\
		& 
		Agglomerative &         0.9811  &         0.1146  & \textbf{0.2052} & \textbf{0.8952} & \textbf{0.5000} \\
		\bottomrule
	\end{tabular}
\end{table*}

\subsection{Karel Programming}
\label{sec:Karel-Programming}

Karel, an educational programming language for beginners~\cite{pattisKarelRobotGentle1981}, has been utilized as the testbed by some recent research work in deep learning for neural program synthesis~\cite{devlinNeuralProgramMetaInduction2017,bunelLeveragingGrammarReinforcement2018,chenExecutionGuidedNeuralProgram2018,shinImprovingNeuralProgram2018}. 
It features a ``robot'' living in a grid-world who can move forward, turn left or right, and pick up or put down markers. 
The grammar of the Karel language is shown in \cref{fig:karel-grammar}. 
This is obviously a structured programming language with the control flow constructs sequence, selection and iteration. 
In this paper, we focus on discovering a Karel program's control flow structure from a small number of extraction traces, but leave the logical conditions (for selections or iterations) to future work (see \cref{sec:Future-Work}).
The \texttt{while} and \texttt{repeat} loops are both mapped to the iteration operators ($+$ or $*$) of the structured program process model as defined in \cref{sec:Structured-Program}.

\begin{figure*}[!tb]
	\centering
	\begin{minipage}{\textwidth}
	\begin{framed}
	\begin{IEEEeqnarray*}{rCl}
	\text{Prog }   p & := & \verb|def run|(): s \\
	\text{Stmt }   s & := & a \mid 
	                        s_1 ; s_2 \mid
	                        \verb|if |(b): s \mid
	                        \verb|if |(b): s_1 \verb| else|: s_2 \mid
	                        \verb|while |(b): s \mid 
	                        \verb|repeat |(r): s \\
	\text{Cond }   b & := & \verb|frontIsClear|() \mid
	                        \verb|leftIsClear|()  \mid
	                        \verb|rightIsClear|() \mid
	                        \verb|markersPresent|() \mid \\
	                 &    & \verb|noMarkersPresent|() \mid
	                        \verb|not | b \\
	\text{Action } a & := & \verb|move|() \mid 
	                        \verb|turnRight|() \mid 
	                        \verb|turnLeft|() \mid
	                        \verb|pickMarker|() \mid 
	                        \verb|putMarker|() \\
	\text{Cste }   r & := & 0 \mid 1 \mid \dots \mid 19
	\end{IEEEeqnarray*}
	\end{framed}
	\end{minipage}
	\caption{The domain-specific language for Karel programs \cite{bunelLeveragingGrammarReinforcement2018}.}
	\label{fig:karel-grammar}
\end{figure*}

The Karel programming dataset\footnote{\url{https://msr-redmond.github.io/karel-dataset/}} is a large dataset of simple Karel programs used for training and testing the models synthesizing Karel programs from input-output examples.
We adapt this dataset for synthesizing Karel programs from execution traces instead.
Each Karel program in the dataset comes with six input-output examples. 
For our experiments, we compute the six execution traces (i.e., the sequence of actions) for each Karel program with respect to those six input-output examples, and then filter out the Karel programs which have less than six distinct execution traces.
Thus, we obtain a large set of event logs where each event log contains six traces (cases) generated by a ground-truth Karel program. 
Furthermore, we split the set of event logs into two subsets according to whether the corresponding ground-truth Karel program has duplicate activities or not.
This is to facilitate the investigation of how important it is to accommodate duplicate activities in the process model.
\cref{tab:karel-dataset} shows the descriptive statistics of the \emph{filtered} Karel programming dataset.
For each subset, we have calculated the average length of execution traces, the average number of ground-truth program tokens and the average depth of ground-truth abstract syntax trees.

\begin{table*}[!tb]
	\caption{Descriptive statistics of the filtered Karel programming dataset (where each log has 6 traces).}
	\label{tab:karel-dataset}
	\centering
	\begin{tabular}{@{}l|r|rrr@{}}
		\toprule
		Data Subset & \#Logs & Trace-Length & Prog-Tokens & Tree-Depth \\
		\midrule
		None-Duplicate &  9,088 & 12.32$\pm$09.50  & 5.44$\pm$1.65  & 3.95$\pm$0.79 \\
		With-Duplicate & 25,714 & 18.93$\pm$13.57 & 10.32$\pm$4.19 & 4.35$\pm$0.96 \\
		\bottomrule
	\end{tabular}
\end{table*}

To the best of our knowledge, the inductive miner is the only existing process discovery algorithm that can produce structured programs.
Therefore only the inductive miner is included as the baseline in our experiments on the Karel programming dataset.

As shown in \cref{tab:karel-pd-results}, our proposed agglomerative miner significantly outperforms the inductive miner on both subsets in terms of the standard process discovery performance metrics $F_1$-score, generalization and simplicity.

\begin{table*}[!tb]
	\caption{Process discovery performances on the filtered Karel programming dataset.}
	\label{tab:karel-pd-results}
	\centering
	\begin{tabular}{@{}l|l|ccccc@{}}
		\toprule
		Data Subset & Method & Fitness & Precision & $F_1$-score & Generalization & Simplicity \\
		\midrule
		\multirow{2}{*}{None-Duplicate} & 
		Inductive     &         0.9959  &         0.5338  &         0.6832  &         0.6101  &         0.8264  \\
		& 
		Agglomerative &         0.9903  &         0.7582  & \textbf{0.8433} & \textbf{0.6854} & \textbf{0.8969}  \\
		\hline
		\multirow{2}{*}{With-Duplicate} & 
		Inductive     &         0.9888  &         0.4146  &         0.5717  &         0.6359  &         0.7837  \\
		& 
		Agglomerative &         0.9789  &         0.5209  & \textbf{0.6498} & \textbf{0.6925} & \textbf{0.8027}  \\
		\bottomrule
	\end{tabular}
\end{table*}

More importantly, we propose to measure the performance of structured program synthesis by comparing the structured program (process model) generated by a process discovery algorithm with the ground-truth.   
One metric is the proportion of \emph{exact matches}, i.e., what percentage of generated programs are exactly identical to the true underlying programs.
Since the order of different branches in the selection control flow construct should not affect its semantics, i.e., $(S_1|S_2)$ is equivalent to $(S_2|S_1)$, we sort the branches in all the \texttt{if-then-else} statements beforehand to disregard such superficial differences. 
Another metric is the Levenshtein \emph{edit distance} between each generated program and its corresponding ground-truth program. 
Here we consider each program as a sequence of program tokens rather than a string of characters, so an edit means an insertion, deletion or substitution of not a single character but a single program token.
The smaller the edit distance, the better the generated program, as it is closer to the ground-truth.
Note that both of the above two metrics measure the \emph{syntactic} similarity/discrepancy between programs, which is an underestimation of the effectiveness for program synthesis: it is very possible for two syntacticly different programs to be semantically equivalent (known as \emph{program aliasing}~\cite{bunelLeveragingGrammarReinforcement2018}).
Nevertheless, these two syntactic metrics are obviously still informative and useful.

As shown in \cref{tab:karel-ps-results}, our proposed agglomerative miner works significantly better than the inductive miner for the Karel program synthesis task in terms of both exact matches and edit distances.
If the ground-truth program does not contain duplicate statements (activities), the agglomerative miner can recover it exactly from six traces with a good ($>55\%$) chance, which is about ten times higher than the inductive miner baseline.
Even when the ground-truth program contains duplicate statements (activities), the agglomerative miner still manages to get 1\% exact matches, thanks to the graph rewriting rule \cref{fig:iteration3}. 

\begin{table*}[!tb]
	\caption{Program synthesis performances on the filtered Karel programming dataset.}
	\label{tab:karel-ps-results}
	\centering
	\begin{tabular}{@{}l|l|r@{ = }rr@{}}
		\toprule
		Data Subset & Method & \multicolumn{2}{c}{Exact-Match} & Edit-Dist \\
		\midrule
		\multirow{2}{*}{None-Duplicate} & 
		Inductive     &  517/9088 &           5.69\% &           2.94$\pm$1.80  \\
		& 
		Agglomerative & 5123/9088 & \textbf{56.37}\% &  \textbf{1.24}$\pm$1.73  \\
		\hline
		\multirow{2}{*}{With-Duplicate} & 
		Inductive     &   0/25714 &           0.00\% &           7.18$\pm$3.20  \\
		& 
		Agglomerative & 258/25714 &  \textbf{1.00}\% &  \textbf{5.98}$\pm$3.61  \\
		\bottomrule
	\end{tabular}
\end{table*}

\section{Future Work}
\label{sec:Future-Work}

The agglomerative process discovery algorithm needs to be extended to address the \emph{infrequency} and \emph{incompleteness} of behavior, i.e., the activities that are rarely observed and thus tend to be outliers as well as the activities that have not been recorded in the event log.
In principle, similar techniques from the inductive miner~\cite{leemansDiscoveringBlockStructuredProcess2013a,leemansDiscoveringBlockStructuredProcess2014,leemansScalableProcessDiscovery2015} could be utilized.

This paper has focused on the discovery/synthesis of a program's control flow structure only, but ignored the inference of logical conditions for selection and iteration. 
It is possible to derive such logical conditions by analyzing the states of the environment at and before the point the process branches into different paths according to the recorded execution traces. 
For Karel programs, the state at any moment could be fully specified by four Boolean variables: 
\texttt{frontIsClear}, \texttt{leftIsClear}, \texttt{rightIsClear} and \texttt{markersPresent} (see \cref{fig:karel-grammar}).
The \emph{decision tree} learning algorithm is promising to address this problem, as shown by previous studies~\cite{rozinatDecisionMiningProM2006,shragaInductiveContextawareProcess2019}.

When we evaluate process discovery algorithms for their effectiveness in program synthesis, we have only measured the syntactic equivalence between the generated program and the ground-truth program. 
Ideally, we want to measure the semantic equivalence: whether the two given programs would exhibit identical behavior, i.e., always produce the same output for the same input. 
This metric is partially reflected by the previously mentioned generalization score for process discovery, but a more accurate way to estimate it is to execute two given programs under a large number of conditions and compare their outputs.

The program synthesis experimental results in \cref{sec:Karel-Programming} suggest that duplicate statements (activities) are common in real-world structured programs (process models) but they are not well addressed by existing process discovery algorithms or the current version of agglomerative miner.
This seems to be an important and challenging research problem in the direction towards a unified theory of process discovery and program synthesis.







\section{Conclusion}
\label{sec:Conclusion}

The main contributions of this paper are as follows.
\begin{itemize}
	\item First, we re-examine process discovery from the perspective of program synthesis, and argue that using structured programs directly as target process models would make the translation from abstract process models to executable processes easier to understand and implement, particularly in the context of robotic process automation.
	\item Second, we design an agglomerative process discovery algorithm for structured programs based on iterative graph rewriting, inspired by the popular inductive miner.   
	\item Third, we introduce an evaluation framework for measuring the program synthesis performance of different process discovery algorithms, and demonstrate the advantages of our proposed agglomerative approach over existing methods.
\end{itemize}

\bibliographystyle{abbrv}
\bibliography{../process-mining,../program-synthesis}

\end{document}